\documentclass[11pt,a4paper]{article}

\usepackage[utf8]{inputenc}
\usepackage[T1]{fontenc}
\usepackage[english]{babel}
\usepackage{amsmath,amssymb}
\usepackage{hyperref}
\usepackage{url}
\usepackage{graphicx}
\usepackage{microtype}
\usepackage{textcomp}
\usepackage{booktabs}
\usepackage[margin=2.5cm]{geometry}
\hypersetup{
    pdftitle={On the missing data layer and a potential solution},
    pdfauthor={Francis F Daniel, Mauro Ibañez, Francis Perelman, Marian Basti},
    pdfkeywords={AI, Data Infrastructure, Latin America, Dataset Discovery, Multipolar AI},
    colorlinks=true,
    linkcolor=blue,
    citecolor=blue,
    urlcolor=blue
}

\title{On the missing data layer and a potential solution}
\author{
  Francis F Daniel\textsuperscript{1} \and
  Mauro Ibañez\textsuperscript{1} \and
  Francis Perelman\textsuperscript{2} \and
  Marian Basti\textsuperscript{1}
}
\date{July 29, 2026}

\begin{document}
\maketitle

\begin{center}
\textsuperscript{1}SURUS \quad \textsuperscript{2}Independent\\[0.3em]
Contact: \texttt{francis@surus.lat}
\end{center}

\begin{abstract}
Latin America is missing two foundational layers required to build its own AI: a \textbf{datasets layer} and a \textbf{benchmarks layer}. These are not optional infrastructure --- they are the minimum prerequisites for both building and evaluating AI systems. Their absence carries a dual cost. On the industry side, most AI tasks are universal but the data needed to get state-of-the-art performance in a given context is regionally peculiar. On the public-institution side, the cost is different in kind: models trained elsewhere encode foreign value systems, languages, and historical narratives \cite{tao2024cultural}, shaping how citizens are represented, served, and governed. Together, the absence of both layers constitutes a loss in both industrial productivity and sovereign capacity over a technology that is becoming critical infrastructure. The dataset layer faces two compounding problems: discovery and supply. Latin American datasets exist but are scattered across places like Hugging Face and the appendices of research papers, with no shared index. Even pooled, their volume falls far below what frontier AI development requires. The two problems reinforce each other: with nowhere to publish for return, fewer datasets get released; with few releases, no critical mass forms. The proposed DataHub is designed to break this loop from both ends --- indexing what exists and making contribution a visible, rewarded act. The DataHub is organized around a task-first ontology with three nested levels: \textbf{task} (e.g. transcription), then \textbf{domain} (e.g. medical), then \textbf{language and regional variant} ----- (e.g. Rioplatense Spanish). Each level compounds the value of the last, paying off on both sides: for industry, in performance on the target task; for public institutions and the populations they serve, in representation and fit. We prefer a \textbf{multipolar AI world} and believe Latin America should be one of those poles, contributing on its own terms. The DataHub must therefore be open by design --- because collaboration matters most for regions not at the frontier --- and incentive-driven by construction, since a commons only stays alive when contributors are rewarded. Open questions remain: regulatory heterogeneity across jurisdictions, shared standards for metadata and licensing, collaboration norms, and the institutional incentives needed to unlock data currently kept internal. This paper is a kickstart by the SURUS team on the missing data layer; and an invitation to companies, institutions, and individuals across the region to shape what the foundational infrastructure for Latin American AI should become, and to contribute and benefit from it. We prefer a multipolar Arica should be one ofthose poles, contributing on its own terms. The Hub must therefore be open by design --- because collabons not at the frontier ---and incentive-driven by construction, since a commons only stays alive when contributors are rewarded. Open questions remain: regulatory heterogeneity across jurisdictions, shared standards for metadata and licensing, collaboration norms, and needed to unlock datacurrently kept internal. This paper is a kickstart by the SURUS team --- a working artifact targeting the dataset layer; the benchmark layer is the subject of LatamBoard, athey are an invitation toresearchers, institutionegion to shape what thefoundational infrastructure for Latin American AI should become.
\end{abstract}


\tableofcontents
\newpage

\textit{SURUS Team · Buenos Aires · 2026}

\section{Introduction}

The state of AI in Latin America today is one of consumption. The region uses models built elsewhere, trained on data collected elsewhere, evaluated against tasks defined elsewhere \cite{salaspilco2022latam}. For genuinely universal and verifiable capabilities --- coding, general-purpose reasoning, mathematics --- this is workable; outside those, performance degrades on tasks where regional context (represented through data) matters, and no evaluation relevant to the region exists to measure how much.

Two foundational layers are missing: the \textbf{dataset layer}, which training requires, and the \textbf{benchmark layer}, which evaluation requires. Without datasets, the region cannot train its own models, agents, or downstream systems. Without benchmarks, it cannot evaluate anything it builds --- nor evaluate the foreign systems already in use.

This paper targets the dataset layer; the benchmark layer is addressed in LatamBoard, a companion effort by the SURUS team. It covers the cost of both missing layers, the current state of Latin American datasets, the Hub's task-first ontology, the proposed open design, the geopolitical position from which the design follows, and the open problems that remain.

\section{Two missing layers, two lanes of cost}

These two layers are not optional infrastructure --- they are the \textbf{minimum prerequisites} for both building and evaluating AI systems. Without datasets, the region cannot train models, agents, or downstream systems --- any of them. Without benchmarks, it cannot evaluate what it builds, and --- just as critically --- cannot evaluate the foreign systems already deployed across its institutions and industries. Both absences are active today.

Without a benchmark layer, industries cannot evaluate the models they procure --- and adopted models can underperform silently on the very tasks they were bought to perform. Regional peculiarities matter for most industrially useful tasks; without benchmarks on those tasks, procurement decisions are made blind. The failure is silent: the model returns an output, but not the performance needed for full adoption and diffusion on the tasks that matter \cite{manvi2024geobias}.

For public institutions, the absence of benchmarks forecloses something larger: the possibility of becoming \textbf{independent auditors} of both regional and worldwide AI systems in use. Some universities independently measure poverty or inequality, producing formal, region-anchored evaluations that their countries and the region use. Latin America's universities and public institutions in general could produce the same for AI: formal evaluations of what models --- local and global --- get right and wrong for the populations they serve.

On the industry side, the problem is performance: the task is often universal, but the data needed to perform it at the state-of-the-art is \textbf{regionally peculiar}. A pest-classification model trained on European or North American agriculture will not reliably identify \cite{balasubramaniam2021pest} the pests that determine harvests in Argentina or Brazil. The task is universal; the crops, the species, and the visual distribution are not. This is the performance gap in concrete form.

On the public-institution side, the cost is different in kind: models trained elsewhere encode foreign value systems, languages, and historical narratives --- shaping how citizens are represented, served, and governed. The failure mode is silent --- the model works, returns plausible output, and is simply miscalibrated for the population it serves. The cost is not just performance but inclusion and representation: whether AI systems reflect and serve the societies that use them \cite{bender2021stochastic}.

The two lanes are not independent. The costs compound each other --- and together they constitute a \textbf{loss of sovereign capacity} over a technology that is becoming critical infrastructure \cite{roberts2024sovereignty}. A region that cannot build or evaluate AI for its own industries, languages, and institutions cannot capture the gains this technology offers, nor exercise meaningful agency over how it is used inside its borders.

\section{State of Latin American datasets}

The dataset layer faces two compounding problems that must be addressed together. The first is \textbf{discovery} \cite{nargesian2021datalake}; the second is \textbf{supply}. They reinforce each other --- which is why both must be in scope.

Latin American AI datasets exist --- the problem is that they are scattered across places like Hugging Face \cite{wolf2019transformers}, the Mozilla Foundation, and the appendices of research papers, with no shared index. A practitioner asking \textit{which datasets can I use to train a model for medical transcription in Rioplatense Spanish?} can answer only by personal network or weeks of search. This is not a failure of any platform; it is the absence of a regionally relevant index.

Even with perfect indexing, the total volume of Latin American datasets would remain far below what frontier AI development requires --- and far below what North America, Europe, and Asia have accumulated. This is a structural gap, not a capability one. The gap compounds annually --- every year without an active creation effort is a year the distance grows \cite{mandal2021diversity}.

These two problems reinforce each other in a loop that, left alone, would persist indefinitely. There is no place to publish a Latin American AI dataset where the contributor sees a return on the act of contributing, so fewer datasets get published; because few are published, no critical mass forms to motivate the next contributor \cite{zuiderwijk2020sharing}. The loop closes. The \textbf{DataHub} is designed to break it from both ends --- indexing what already exists (discovery) and making contribution a visible, rewarded act (supply).

\section{A task-first ontology for AI}
Existing dataset catalogs almost universally organize entries by input modality or by target model --- we believe both are wrong. An AI model is fundamentally a program for a task, and task is therefore the first axis. Tasks include transcription, classification, extraction, translation, forecasting, segmentation, and others.

Existing dataset catalogs almost universally organize entries by input modality or by target model --- we believe both are wrong for AI, because an AI model is fundamentally a \textbf{program} that performs a \textbf{task} \cite{akhtar2024croissant}. This is also in line with how an AI practitioner works: he/she doesn't start from \textit{I have an image} or \textit{I want to use model X} --- they start from \textit{I need a system that does Y.} \cite{paullada2021data} Thus the right ontology is one organized around what an AI system \textit{does} (e.g. extraction, classification, reasoning, transcription, etc.). This DataHub adopts a \textbf{task-first ontology} with two nested levels: \textbf{domain} \& \textbf{language}. This is represented as \texttt{/<task?>/<domain?>/<language?>}.

Domain is the second axis because a task without domain context is almost always too broad to achieve the needed performance. Within transcription, medical and legal are different datasets --- vocabulary, annotation conventions, etc. A model trained on medical transcription underperforms on legal-domain audio files even within the same language.

The third axis is language --- and within language, \textbf{regional variant}. Medical transcription in Rioplatense Spanish is not interchangeable with Colombian Spanish or Brazilian Portuguese \cite{goncalves2015dialects}: clinical vocabulary, patient-facing register, and acoustic patterns all differ.

Each level compounds the value of the last: general transcription is useful, medical transcription is more useful, medical transcription in Rioplatense Spanish is more useful still. The compounding pays off on both sides. For industry, each layer of specificity buys a performance dividend --- a Buenos Aires hospital deploying a medical-transcription model tuned to Rioplatense Spanish gets a performance a general model cannot deliver. For public institutions and the populations they serve, the same specificity buys representation and inclusivity --- models that actually work for the people who use them.

A practitioner walks this hierarchy from general to specific, and the ontology is designed so that walk is the natural one. The same structure guides a contributor: declaring task, domain, and variant places a dataset in the exact slot where someone searching for it will look. The ontology is the search interface and the publishing form in one.

\section{Open by design, incentive-driven by construction}

On the geopolitical side, we prefer a \textbf{multipolar AI world}, and we believe Latin America should be one of those poles --- with its own technology and its own agency. This is a position against both single-pole dominance --- a world where AI is defined by the United States or China alone --- and against a few-pole world that excludes Latin America. The more regions that develop their own AI systems, the better; this is the region's stake in that outcome. It is not an argument against collaboration with other regions; it is an argument \textit{for} Latin America contributing to that collaboration on its own terms.

The question of how to build it --- open or closed, shared or proprietary --- is a strategic choice, not a neutral technical decision. Different regions have taken different routes: some have built largely closed and proprietary AI ecosystems; others have built open and collaborative ones \cite{corralesgaray2024cocreating}. Both can produce AI capability. The choice shapes who can participate and how fast a non-frontier region can close the gap.

We argue \textbf{open by design} is right for Latin America for a region-specific reason: since collaboration fosters development, collaboration matters most for regions not at the frontier of a technology \cite{lernertirole2002opensource}. For a non-frontier region, the priority is not to protect what it has from competition but to contribute as much as possible, compounding collaborative efforts to close the gap.

A commons is not self-sustaining by virtue of being a commons \cite{ostrom2010polycentric}. Without explicit incentives, there's no intertia to continue publishing data into a hub. Each new dataset compounds the value of every prior contribution --- the index grows more complete, the ontology more populated. Contributors gain something concrete: visibility within the regional AI community, attribution attached to their dataset, and recognition for their institution through brand-awareness.

The DataHub is not designed as a static catalog but as a \textbf{continuum}: it has active contribution, curation, and growth. The contribution path must be low-friction --- a researcher who has just produced a dataset should be able to publish it in minutes \cite{borgman2012conundrum}.

\section{Open problems}

The DataHub does not resolve every question implied by the dataset layer it begins to build.

Data protection, copyright, and public-data access regimes differ country by country across Latin America. The Hub cannot resolve this --- these are matters of national law. What it can and must do is surface the licensing and legal-jurisdiction information of every indexed dataset \cite{longpre2024licensing}, so practitioners decide with the relevant context visible \cite{edwards2011friction}.

There are no shared standards for dataset metadata, licensing, or quality in the region \cite{gebru2021datasheets}. Two datasets nominally about \textit{Spanish transcription} may use incompatible annotation schemas, incompatible licenses, and incompatible quality bars \cite{musen2022metadata} Collaboration norms --- attribution, versioning, and arbitration when a community contests how it is represented --- also remain undefined. When one party improves another's dataset, how is credit allocated? These are questions for the regional community to decide together, not for one team to impose.

Companies, universities, governments, and public bodies across Latin America hold significant volumes of data that would be valuable as published datasets --- and most of it does not get published. The right incentives for each of these to publish their data remain unknown \cite{tenopir2011sharing}. The DataHub's visibility mechanisms address part of this --- publishing now produces recognition. But the deeper problem requires policy work outside the DataHub: grant requirements that mandate dataset publication, university promotion criteria that value dataset contributions, government open-data policies that treat AI-readable formats as a deliverable.

\section{A kickstart and an invitation}

This paper is a \textbf{kickstart} --- a working artifact and a position targeting the dataset layer specifically; the benchmark layer is the subject of LatamBoard, a companion effort. The SURUS team built the first version of the Data Hub because somebody had to begin. Together, the Data Hub and LatamBoard are an invitation --- not a finished proposal --- and not the work of a single team.

To companies, researchers and institutions holding data: we invite you to publish it into the Hub. Whether the dataset is a paper appendix from three years ago, an active working corpus, or an internal archive that has never been released --- the DataHub is built to receive it. Each contribution compounds the value of every prior one.

To AI practitioners across the region: we invite you to use the DataHub and contribute to its improvement. The most important signal that matters is practitioners choosing datasets from it, building models with it, and contributing datasets to it. Usage and contribution is what converts a commons from a directory into infrastructure --- and feedback from use is what improves it.

To the broader community --- universities, foundations, governments, companies: help shape what the foundational infrastructure for AI in Latin America should become, contribute, and then benefit from it. The ontology will evolve, standards must be debated, and collaboration norms must be defined. None of this is one team's work --- and none of the benefit accrues to a single organization.

The dataset layer for Latin American AI will be built by the region, for the region, or it will not be built at all. The DataHub is one step toward that future, and we invite you to join us.

\textbf{Availability.} The Data Hub is live at [https://datahub.lat](https://datahub.lat).

\bibliographystyle{plain}
\bibliography{main}

@article{roberts2024sovereignty,
  author = {Roberts, Huw},
  title = {Digital sovereignty and artificial intelligence: a normative approach},
  journal = {AI and Ethics},
  year = {2024},
  doi = {10.1007/s10676-024-09810-5}
}

@article{goncalves2015dialects,
  author = {Gon{\c{c}}alves, Bruno and S{\'a}nchez, David},
  title = {Learning about Spanish dialects through Twitter},
  journal = {arXiv preprint arXiv:1511.04970},
  year = {2015}
}

@inproceedings{manvi2024geobias,
  author = {Manvi, Raghav and Khanna, Saachi and Burke, Marshall and Lobell, David B. and Ermon, Stefano},
  title = {Large Language Models are Geographically Biased},
  booktitle = {ICML 2024},
  year = {2024},
  note = {arXiv:2402.02680}
}

@inproceedings{bender2021stochastic,
  author = {Bender, Emily M. and Gebru, Timnit and McMillan-Major, Angelina and Shmitchell, Shmargi},
  title = {On the Dangers of Stochastic Parrots: Can Language Models Be Too Big?},
  booktitle = {Proc. 2021 ACM FAccT},
  year = {2021},
  doi = {10.1145/3442188.3445922}
}

@article{lernertirole2002opensource,
  author = {Lerner, Josh and Tirole, Jean},
  title = {Some Simple Economics of Open Source},
  journal = {The Journal of Industrial Economics},
  volume = {50},
  number = {2},
  pages = {197--234},
  year = {2002},
  doi = {10.1111/1467-6451.00174}
}

@article{wolf2019transformers,
  author = {Wolf, Thomas and others},
  title = {HuggingFace's Transformers: State-of-the-art Natural Language Processing},
  journal = {arXiv preprint arXiv:1910.03771},
  year = {2019}
}

@article{tao2024cultural,
  author = {Tao, Yan and others},
  title = {Cultural bias and cultural alignment of large language models},
  journal = {PNAS Nexus},
  year = {2024},
  doi = {10.1093/pnasnexus/pgae346}
}

@article{gebru2021datasheets,
  author = {Gebru, Timnit and others},
  title = {Datasheets for Datasets},
  journal = {Communications of the ACM},
  year = {2021},
  doi = {10.1145/3458723}
}

@article{longpre2024licensing,
  author = {Longpre, Shayne and others},
  title = {A large-scale audit of dataset licensing and attribution in AI},
  journal = {Nature Machine Intelligence},
  year = {2024},
  doi = {10.1038/s42256-024-00878-8}
}

@article{ostrom2010polycentric,
  author = {Ostrom, Elinor},
  title = {Beyond Markets and States: Polycentric Governance of Complex Economic Systems},
  journal = {American Economic Review},
  volume = {100},
  number = {3},
  pages = {641--672},
  year = {2010},
  doi = {10.1257/aer.100.3.641}
}

@article{nargesian2021datalake,
  author = {Nargesian, Fatemeh and others},
  title = {Data lake management},
  journal = {Proceedings of the VLDB Endowment},
  year = {2021},
  doi = {10.14778/3352063.3352116}
}

@article{balasubramaniam2021pest,
  author = {Balasubramaniam, Sasi and others},
  title = {Machine Learning based Disease and Pest detection in Agricultural Crops},
  journal = {EAI Endorsed Transactions on Internet of Things},
  year = {2021},
  doi = {10.4108/eetiot.5049}
}

@article{paullada2021data,
  author = {Paullada, Amandalynne and Raji, Inioluwa Deborah and Bender, Emily M. and Denton, Emily and Hanna, Alex},
  title = {Data and its (dis)contents: A survey of dataset development and use in machine learning research},
  journal = {Patterns},
  year = {2021},
  doi = {10.1016/j.patter.2021.100336}
}

@article{salaspilco2022latam,
  author = {Salas-Pilco, Sergio Z. and Yang, Yang},
  title = {Artificial intelligence applications in Latin American higher education: a systematic review},
  journal = {Intl J. of Educational Technology in Higher Education},
  year = {2022},
  doi = {10.1186/s41239-022-00326-w}
}

@inproceedings{mandal2021diversity,
  author = {Mandal, Aprameya and Leavy, Saoirse and Little, Suzanne},
  title = {Dataset Diversity},
  booktitle = {Proc. 1st Intl Workshop on Trustworthy AI for Multimedia Computing},
  year = {2021},
  doi = {10.1145/3475731.3484956}
}

@article{zuiderwijk2020sharing,
  author = {Zuiderwijk, Anneke and Shinde, Rishabh and Jeng, Wei},
  title = {What drives and inhibits researchers to share and use open research data?},
  journal = {PLOS ONE},
  year = {2020},
  doi = {10.1371/journal.pone.0239283}
}

@inproceedings{akhtar2024croissant,
  author = {Akhtar, Mubashara and others},
  title = {Croissant: A Metadata Format for ML-Ready Datasets},
  booktitle = {Proc. Eighth Workshop on Data Management for End-to-End ML},
  year = {2024},
  doi = {10.1145/3650203.3663326}
}

@article{corralesgaray2024cocreating,
  author = {Corrales-Garay, David and Rodr{\'i}guez-S{\'a}nchez, Jos{\'e} L. and Montero-Navarro, Andr{\'e}s},
  title = {Co-Creating Value With AI: A Bibliometric Approach to the Use of AI in Open Innovation Ecosystems},
  journal = {IEEE Access},
  year = {2024},
  doi = {10.1109/ACCESS.2024.3391054}
}

@article{borgman2012conundrum,
  author = {Borgman, Christine L.},
  title = {The conundrum of sharing research data},
  journal = {J. of the American Society for Information Science and Technology},
  year = {2012},
  doi = {10.1002/asi.22634}
}

@article{edwards2011friction,
  author = {Edwards, Paul N. and Mayernik, Matthew S. and Batcheller, Archer L. and Bowker, Geoffrey C. and Borgman, Christine L.},
  title = {Science friction: Data, metadata, and collaboration},
  journal = {Social Studies of Science},
  year = {2011},
  doi = {10.1177/0306312711413314}
}

@article{musen2022metadata,
  author = {Musen, Mark A. and others},
  title = {Modeling community standards for metadata as templates makes data FAIR},
  journal = {Scientific Data},
  year = {2022},
  doi = {10.1038/s41597-022-01815-3}
}

@article{tenopir2011sharing,
  author = {Tenopir, Carol and others},
  title = {Data Sharing by Scientists: Practices and Perceptions},
  journal = {PLoS ONE},
  year = {2011},
  doi = {10.1371/journal.pone.0021101}
}

\end{document}